\documentclass[11pt]{article}

\usepackage[final]{acl}
\usepackage{times}
\usepackage{latexsym}
\usepackage[T1]{fontenc}
\usepackage[utf8]{inputenc}
\usepackage{microtype}
\usepackage{inconsolata}
\usepackage{graphicx}
\usepackage{booktabs}
\usepackage{amsmath}
\usepackage{multirow}
\usepackage{colortbl}
\usepackage{xcolor}
\usepackage{enumitem}
\usepackage{placeins}
\usepackage{stfloats}

\definecolor{lightgray}{gray}{0.85}

\title{Local and Global Regimes of Geometric Complexity in Language Model Representations}
\author{
  Arwa Osman$^1$, Marco Baroni$^{1,2}$, Iuri Macocco$^1$ \\
  $^1$Universitat Pompeu Fabra (UPF) \\
  $^2$ICREA \\
  \\
  \textbf{Correspondence:} \texttt{arwa.osman@upf.edu}
}

\begin{document}
\maketitle

\begin{abstract}
Intrinsic dimensionality (ID) is widely used to probe the representational 
complexity of language models, but it remains unclear whether ID differences 
reflect properties of language itself or unchecked characteristics of the underlying dataset. In this paper, we focus specifically on how lexical diversity, the number of unique last-token items present in a dataset, affects ID estimates of that dataset. We find a 
scale-dependent transition between two regimes: at low lexical diversity, 
conditions with fewer unique final words produce higher ID, while at high lexical 
diversity, this ordering reverses, and conditions with more unique words produce 
higher ID. We derive an exact, parameter-free formula for the point at which 
this reversal occurs, which matches the 
observed transition point at every scale tested. On the one hand, our results highlight how care must be taken when interpreting the intrinsic dimensionality of a set of representations as a straightforward cue of their complexity. On the other hand, our discovery of the two ID regimes reveals a general principle of organisation of linguistic data in LLMs that sheds new light on their inner manifold structures.
\end{abstract}

\section{Introduction}
\label{sec:introduction}

Intrinsic dimensionality (ID) is the effective number of degrees of freedom needed to describe the structure of high-dimensional data \citep{ansuini2019intrinsic, recanatesi2019, pope2021intrinsic}, and it has become a common tool in NLP for probing the representational complexity of language models \citep{tulchinskii2023intrinsic, valeriani2023geometry, cheng2025emergence}: how ``spread out'' or constrained hidden states are across layers or models.

\begin{figure}[h!]
    \centering
    \includegraphics[width=\columnwidth]{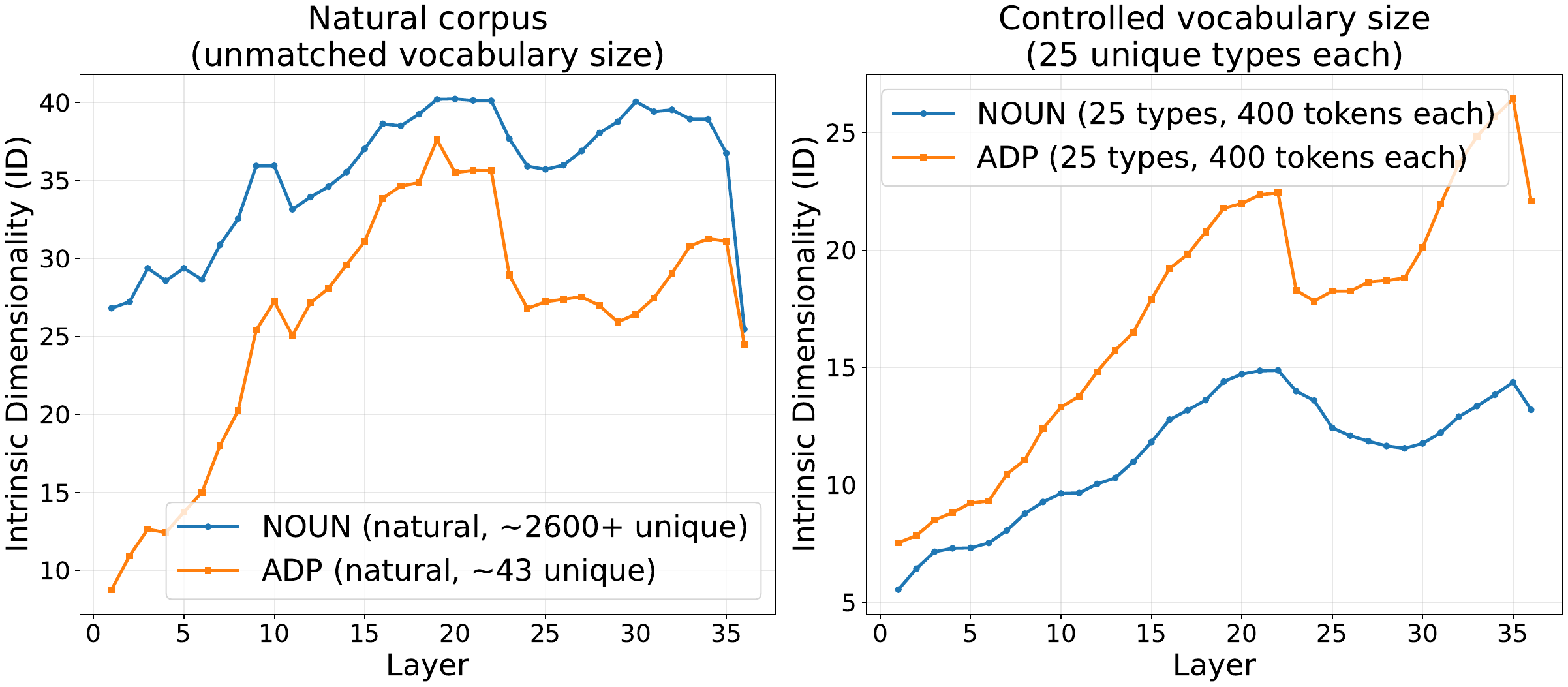}

\caption{Apparent NOUN vs.\ ADP differences in intrinsic dimensionality reverse once lexical diversity is controlled (GRIDE, Qwen3-8B, $k=128$; $N=10{,}000$ samples). \textbf{Left:} naive comparison, unmatched vocabulary size (NOUN: 2600+ types; ADP: $\sim$43 types). \textbf{Right:} both classes matched to 25 types, with 400 tokens per type (Section~\ref{sec:setup}).}
    \label{fig:noun_adp}
\end{figure}

However, it remains poorly understood how ID measurements depend on properties of the underlying dataset. For example, a natural question is whether different parts of speech differ in intrinsic dimensionality \citep{domenichelli2026linguistic}. In a straightforward comparison, they appear to: nouns occupy a substantially higher-ID space than prepositions across nearly every layer of a transformer (Figure~\ref{fig:noun_adp}, left). One could be tempted to conclude that the larger semantic variety of nouns is reflected by richer geometric structure in the transformer representations.

But this comparison is not as straightforward as it first appears. In natural text, the noun class contains thousands of unique word types, while prepositions are drawn from a closed set of only a few dozen. When we repeat the comparison after matching the two classes in both lexical diversity and token frequency, the pattern does not just weaken; it reverses: prepositions now show \textit{higher} ID than nouns across the layers (Figure~\ref{fig:noun_adp}, right). 
What looked like a part-of-speech effect turns out to strongly depend also on the lexical diversity of the samples used to measure ID.

This example leads to the question we address in this paper: independent of syntactic category or semantic content, how does lexical diversity shape the intrinsic dimensionality of contextualised word representations? We isolate lexical diversity as a controlled experimental variable. Using a custom WikiText-derived dataset, we construct eleven conditions spanning from 1 to 10,000 unique contextualised nouns, holding total sample size fixed at $N=10{,}000$ across all conditions, and estimate ID using the generalised ratios intrinsic dimension estimator (GRIDE) \citep{denti2022intrinsic} across the hidden states of eight models from various families and different sizes.

Our results reveal a transition in ID that depends on how the estimator's neighbourhood scale compares to the number of samples available per word: when the scale is small relative to this sample count, ID reflects a word's own contextual variability, a condition we refer to as the ``local'' regime. Differently, once the probing scale is larger than such sample size, the ID instead reflects the diversity of the vocabulary as a whole, moving into a ``global'' regime. We derive the exact point at which this shift occurs and confirm it empirically across models, 
estimators, and word classes.

Any comparison of ID across categories, whether word classes, languages, domains, or models, should take into account that the categories may differ in factors beyond the property under study. We show that differences in lexical diversity can substantially change ID estimates. 
More generally, our results show how the numerical values returned by standard ID estimators might, depending on scale, be based on very different views of the same broader manifold.

\section{Related Work}
\label{sec:related_work}

\paragraph{Methods for estimating intrinsic dimensionality.}
Intrinsic dimensionality (ID) refers to the minimal number of parameters required 
to describe a data representation without significant loss of information 
\citep{ansuini2019intrinsic}, capturing the effective number of degrees of 
freedom along the manifold on which the data resides \citep{pope2021intrinsic}. Estimating ID in high-dimensional data is a well-studied problem, with methods 
developed to address challenges of noise, scale, and computational cost 
\citep{camastra2003data, levina2004maximum}. Maximum Likelihood 
Estimation (MLE), proposed by \citet{levina2004maximum}, models the points 
within a small neighbourhood of a query point as a homogeneous Poisson process, 
and estimates ID from the log-ratio of the distance to the $k$-th nearest 
neighbour relative to each closer neighbour; here, $k$ directly sets the size 
of the local neighbourhood used to estimate the density around each point, with 
larger $k$ incorporating more neighbours into the estimate at the cost of a 
stronger assumption of locally constant density. This estimator is 
computationally efficient but can underestimate ID in high-dimensional or 
complex-manifold settings due to negative bias \citep{denti2022intrinsic}. 
TwoNN \citep{facco2017estimating} instead relies on the ratio of distances to the first and second nearest neighbours, reducing sensitivity to curvature and density variations but remaining sensitive to noise at short scales; a common mitigation, data decimation, reduces this bias at the cost of statistical power \citep{denti2022intrinsic}. GRIDE \citep{denti2022intrinsic} addresses these 
limitations by leveraging higher-order nearest neighbours to estimate ID across 
multiple neighbourhood scales, mitigating noise by modelling distance ratios at 
larger scales without the power loss associated with decimation. This 
multi-scale property is central to our approach, as it allows us to directly 
examine how ID estimates shift with neighbourhood scale $k$.

\paragraph{Intrinsic dimensionality in neural network representations.}

ID has been used to study the geometry of representations in deep networks 
since \citet{ansuini2019intrinsic} showed that ID follows a characteristic 
hump-shaped profile across layers, first expanding, probably to disentangle features in early layers, before compressing toward the output, with lower ID in the final hidden layer correlating with better generalisation. \citet{recanatesi2019} further established that this compression, the degree to which a network 
reduces the dimensionality of its final-layer representations, is itself a 
strong predictor of generalisation to unseen data. \citet{pope2021intrinsic} 
extended this line of work by estimating the ID of common image datasets, 
finding that even high-resolution ImageNet images have an intrinsic 
dimensionality several orders of magnitude smaller than their embedding 
dimension, and showing that dataset ID correlates with learning difficulty. 
More recently, \citet{cheng2025emergence} identified a distinct high-dimensional 
``abstraction phase'' in transformer language models, in which ID rises sharply 
at intermediate layers before collapsing toward lower-dimensional 
representations near the output, interpreted as evidence that models integrate abstract linguistic information before refining it in later layers.

\paragraph{ID as a cue of complexity in NLP settings.}
ID has increasingly been used as a cue of linguistic or representational 
complexity in NLP. \citet{tulchinskii2023intrinsic} used ID to detect 
AI-generated text, exploiting the observation that machine-generated text 
occupies a lower-dimensional subspace than human-written text. 
\citet{yin2024does} characterised model truthfulness in question answering 
using local intrinsic dimension, showing that hallucinated model outputs 
occupy higher-dimensional activation manifolds than correct ground-truth 
answers. \citet{valeriani2023geometry} examined the geometry of hidden 
representations across transformer layers to probe representational complexity more 
broadly. More directly relevant to our concerns, \citet{domenichelli2026linguistic} 
find that open-class words occupy more isotropic, higher-dimensional subspaces 
than closed-class words across both encoder and decoder architectures, and 
note that the two groups also differ in lexical and syntactic diversity, 
without isolating diversity as a controlled variable. Similarly, 
\citet{baroni2026tracing} trace how ID varies across a range of linguistic 
phenomena to draw conclusions about their relative complexity. 

Taken together, the current literature has established ID as a valuable probe of representational complexity in both vision and language models. Yet, input diversity remains an overlooked source of variation in ID estimates. To our knowledge, no prior work isolates lexical diversity as an experimental variable in its own right. We address this gap directly.

\paragraph{Confounds in ID estimation.}
A separate line of work has questioned whether ID estimates on neural representations reliably track the quantity they are assumed to measure. \citet{schulte2026rethinking}, in particular, show both theoretically and empirically that common ID estimators do not track the true underlying ID of a representation, and investigate which factors actually drive commonly reported ID results in the literature.

\section{Experimental Setup}
\label{sec:setup}
\subsection{Dataset}
\paragraph{Sample Construction.}
We use WikiText-103
\citep{merity2017pointer}, a large-scale English 
corpus derived from verified Wikipedia articles, which we 
part-of-speech tag using spaCy \citep{honnibal2020spacy}. From the 
tagged corpus, we construct approximately 84 million samples, each 
consisting of four consecutive non-overlapping sentences, with a 
target word randomly selected from the fourth sentence and its POS 
tag recorded; placing the target in the fourth sentence ensures that three complete sentences of preceding context are available to the model. Each sample is truncated at the target 
word, making it the final token the model processes.

We focus on nouns as target words, since their large natural vocabulary 
allows us to construct conditions spanning a wide range of lexical 
diversity, from a single repeated noun to thousands of unique types. 
To ensure POS purity and minimise tagging noise, we retain only words 
tagged as a noun in at least 80\% of their occurrences and appearing at 
least 300 times in our corpus, yielding a contextualised noun pool 
from which our dataset conditions are constructed. We additionally 
construct a POS-diversified variant of this dataset, sampling target 
words across multiple POS categories rather than nouns only; details 
are given in Appendix~\ref{sec:appendix-allpos}.

\paragraph{Dataset Conditions.}
From the contextualised noun pool, we construct eleven datasets at 
different lexical diversity levels, each fixed at $N = 10{,}000$ 
samples, by sampling $n$ unique nouns with $m = N/n$ samples per 
noun. This is the quantity to be compared with the estimator's neighbourhood scale $k$.
Critically, lexical diversity here refers only to the target noun's 
identity, not its context: the preceding sentences still differ across 
samples regardless of diversity level, drawn independently from the 
corpus, so our conditions isolate the target word's diversity alone. To assess robustness, for each lexical diversity level we construct four independent partitions, each with the same set of unique nouns but different randomly drawn samples. The eleven conditions are summarised in Table~\ref{tab:datasets}.
\begin{table}[h!]
\centering
\footnotesize
\setlength{\tabcolsep}{4pt}
\renewcommand{\arraystretch}{1.1}
\begin{tabular}{rr}
\toprule
\textbf{Unique Nouns} & \textbf{Samples per Noun} \\
\midrule
1      & 10,000 \\
25     & 400 \\
50     & 200 \\
100    & 100 \\
250    & 40 \\
500    & 20 \\
1,000  & 10 \\
1,250  & 8 \\
2,500  & 4 \\
5,000  & 2 \\
10,000 & 1 \\
\bottomrule
\end{tabular}
\caption{Lexical diversity conditions.}
\label{tab:datasets}
\end{table}
\FloatBarrier

\subsection{Models}
We use Qwen3-8B \citep{yang2025qwen3} (36 layers, hidden size 4096) as our primary model. We additionally inspect Meta-Llama-3-8B \citep{meta2024llama3}, Mistral-7B \citep{jiang2023mistral}, DeepSeek-LLM-7B \citep{deepseekai2024deepseekllm}, and four models from the Pythia suite \citep{biderman2023pythia} (410M, 1.4B, 6.9B and 12B parameters) for cross-architecture and model-size validation (Section~\ref{sec:lama-replication}).

\subsection{ID Estimation}
For each sample, we extract the hidden state at the final token position from 
every layer of the model. We then estimate the intrinsic dimensionality of 
these representations using GRIDE. Formally, for a point $i$ with $n_1$-th and 
$n_2$-th nearest-neighbour distances $r_{i,n_1}$ and $r_{i,n_2}$, GRIDE 
estimates the intrinsic dimension $d$ by maximum likelihood from the 
distribution of the ratio $\mu_i = r_{i,n_2}/r_{i,n_1}$, whose distribution depends 
on $d$. We follow the $n_2 = 2n_1$ convention \citep{denti2022intrinsic} as a robust trade-off between scale 
coverage and computational cost, and set $k = n_2$ (so $n_1 = k/2$), referring 
to $k$ as the neighbourhood scale used by GRIDE at each estimate.
Intuitively, $k$ sets how far into the representation space the estimator looks when comparing distances: a small $k$ captures the local geometry immediately surrounding a point, while a larger $k$ captures how that point's neighbourhood relates to increasingly distant regions of representation space. We run GRIDE at six scales 
$k \in \{16, 32, 64, 128, 256, 512\}$.

The computation is performed using the DADApy library \citep{glielmo2022dadapy}. ID is estimated independently for each of the four partitions per condition, and we report the mean and standard deviation of these estimates across partitions. This variance is typically extremely low, so the standard deviation is often not visible in the plots.

\section{A Scale-Dependent Transition in ID}
\label{sec:results}
\begin{figure}[h!]
\includegraphics[width=1\columnwidth]{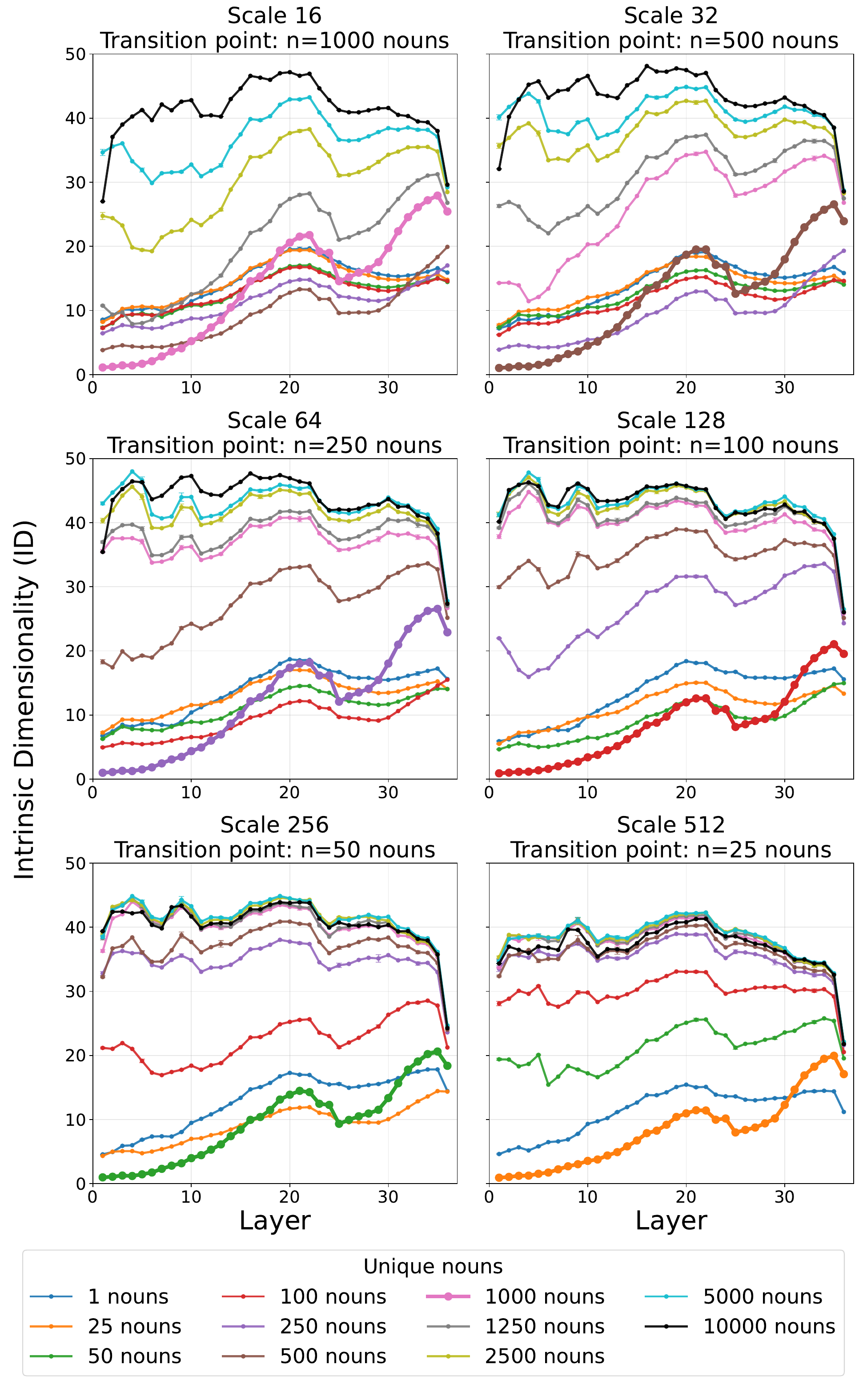}
  \caption{ID curves across layers for all considered scales and lexical-diversity conditions on Qwen3-8B, one panel per scale $k$ and one line per number of unique nouns $n$. The ``transition point'' curve is shown with a thicker line.}
  \label{fig:gride}
\end{figure}

Figure~\ref{fig:gride} shows ID across layers for Qwen3-8B at six scales. At every scale, conditions separate into two regimes that hold across all layers: 
low-diversity conditions remain in a low-ID regime throughout the network, while 
high-diversity conditions remain in a high-ID regime throughout. In the low-ID regime, the relationship between lexical variety and ID is inverse: conditions with \textit{fewer} unique nouns have higher ID. In the high-ID regime, the relationship becomes direct: \textit{more} unique nouns lead to higher ID, up to a level at which the ID estimate tends to stabilise across datasets.

Between these two regimes lies a single condition, a ``transition point'', whose curve (displayed with a thicker line in the figure) does not stay in either regime: it starts near the low-ID cluster in early layers and rises across layers to join the high-ID cluster by the final layers. This transition point shifts systematically with scale: as $k$ increases, it occurs at progressively lower lexical diversity.

A complementary view of the same phenomenon, organised by lexical diversity level rather than scale, is given in Appendix~\ref{sec:appendix-transition-by-n} (Figure~\ref{fig:transition_by_n}).

\FloatBarrier
\FloatBarrier

\subsection{Local and Global Regimes}
\begin{figure}[h!]
\centering
\includegraphics[width=0.56\columnwidth]{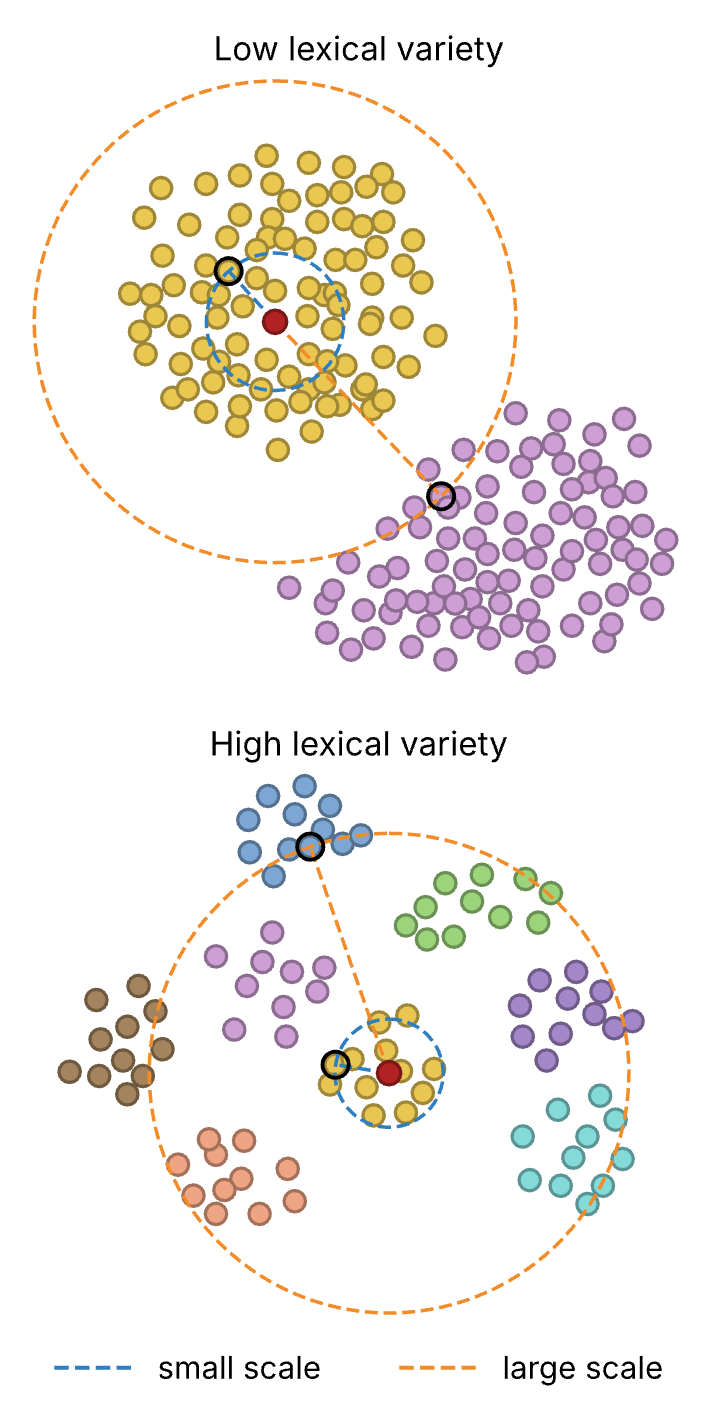}
\caption{Schematic illustration of the local and global regimes. \textbf{Top:} at low lexical diversity, a small neighbourhood scale (blue) stays within the same word's cluster (local regime), while a large scale (orange) already extends into a different word's cluster (global regime). \textbf{Bottom:} at high lexical variety, clusters are smaller and more numerous, so even a small scale can extend beyond the target word's own cluster.}
\label{fig:local_global_schematic}
\end{figure}

We now explain why these two regimes exist, and why the transition between them 
occurs where it does. Since GRIDE is a nearest-neighbour estimator, the 
transition in ID must ultimately be explained by a change in the neighbour structure of the data. We build on prior evidence that, despite being context-sensitive, representations of the same word in different contexts 
remain more similar to one another than representations of different words 
\citep{ethayarajh-2019-contextual}, and assume that this holds strongly enough for 
different contextualised representations of the same word to cluster closer 
together than representations of different words. Given this assumption, a 
neighbourhood of size $k$ falls into one of two cases: either all $k$ 
nearest neighbours belong to the same word type, the local regime, or at least 
one neighbour belongs to a different word type, the global regime (Figure~\ref{fig:local_global_schematic}).

In the local regime, the estimator samples entirely within a single word's 
representation cloud, measuring how the model varies representations of the same 
word across different contexts. In the global regime, the neighbourhood extends 
beyond the word cloud, and the estimator instead measures the geometry of the 
broader noun space. 

\FloatBarrier

\subsection{Deriving the Transition Point}
\label{sec:transition_point}

Let $n$ be the number of unique nouns, $m = N/n$ the number of samples per noun, and $k$ the GRIDE scale. For any query point, there are at most $m-1$ other samples ending with the same word.

GRIDE estimates ID by comparing the distances to two neighbours, the $(k/2)$-th and the $k$-th, so each estimate depends on two neighbourhood scales.

When $k < m$, both neighbours can still be samples of the same word, so the estimate stays in the local regime. When $k \geq m$ but $k/2 < m$, the closer neighbour can still be a same-word sample, but the farther one cannot. The estimate then mixes the two regimes, and this is the transition we observe in our results (the bold lines in Figure~\ref{fig:gride}). When $k/2 \geq m$, both neighbours must come from other words, so the estimate is in the global regime.

The transition therefore begins at $k = m$, the point where the farther neighbour first leaves the same-word cluster:
\begin{equation}
  m = k
  \;\Longrightarrow\;
  \frac{N}{n} = k
  \;\Longrightarrow\;
  n_{\mathrm{transition}} = \frac{N}{k}.
  \label{eq:transition}
\end{equation}

The minimum fraction of the $k$ neighbours that must come from different word types, which we call the \textbf{global ratio}, is:
\begin{equation}
  \rho_{\min}(n,k)
  =
  \max\!\left(
    0,\,
    1-\frac{m-1}{k}
  \right)
  .
  \label{eq:globalratio}
\end{equation}

\subsection{Confirming the Transition Across Scales}
\label{sec:data_collapse}
Our derivation implies that the transition point should depend only on the ratio 
$k/m$, rather than on $k$ and $m$ individually. If this holds, curves 
measured at very different scales ($k=16$ to $k=512$) should collapse onto the 
same trajectory when plotted against this ratio, each entering the 
transition at $k/m=1$ and exiting it at $k/m=2$.

\begin{figure}[h!]
\includegraphics[width=\columnwidth]{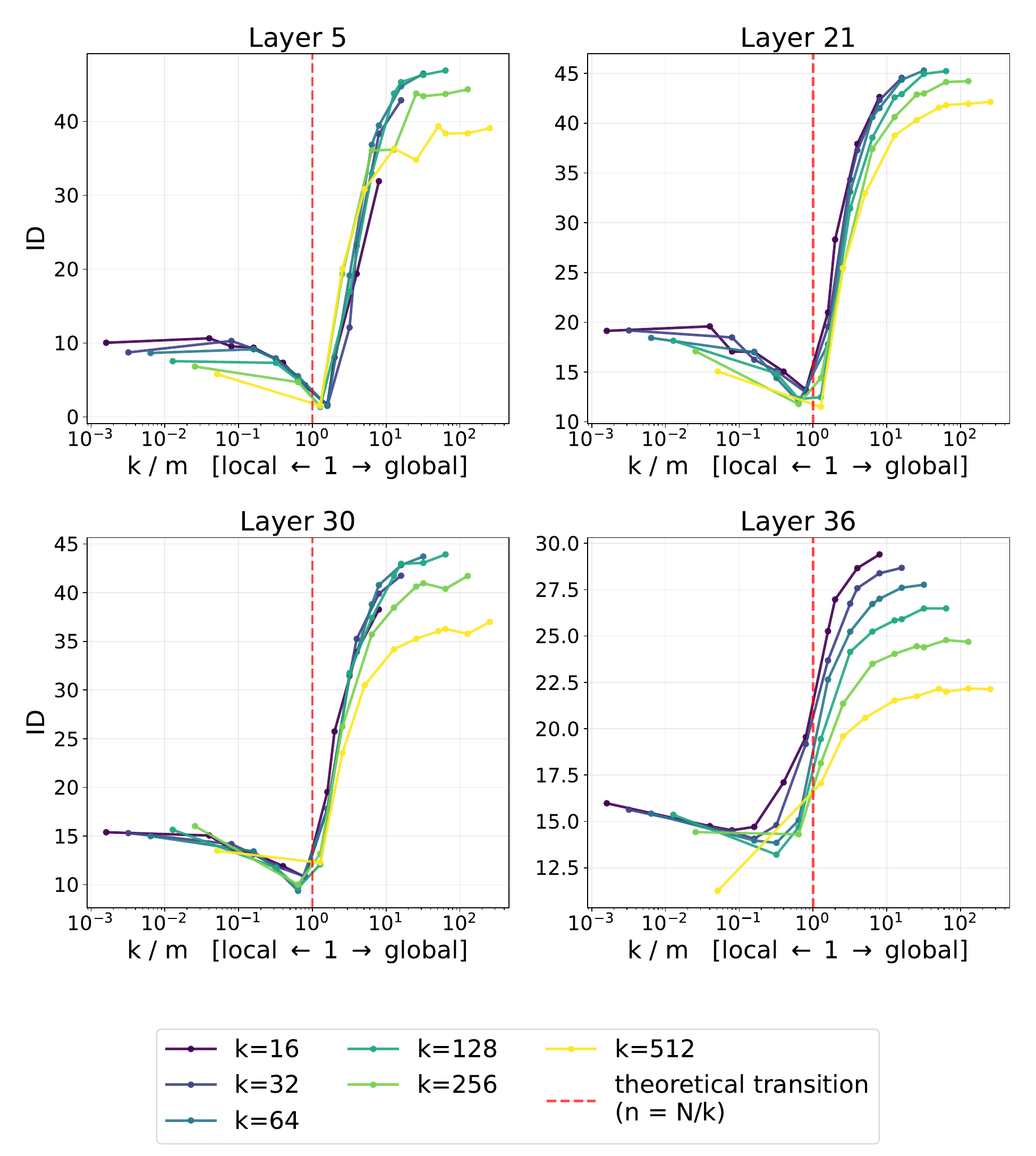}
\caption{ID as a function of $k/m$ across four layers of Qwen3-8B (layers 5, 21, 30, and 36).}  \label{fig:collapse}
\end{figure}

Figure~\ref{fig:collapse} tests this prediction. Each panel shows one layer of 
Qwen3-8B, with one line per scale $k$. Despite this wide range of raw $k$ values, 
all six lines collapse onto a single trajectory once plotted against $k/m$, 
and all of them rise through the predicted transition point (dashed red line), 
with no fitted parameters. This holds consistently across early, middle, and 
late layers (5, 21, 30, and 36), indicating that a single rule, governed 
entirely by the neighbourhood-counting argument above, explains the transition 
regardless of scale or network depth.
We note that layer 36, the 
final layer, shows a somewhat noisier collapse than earlier layers, likely 
reflecting a broader reshuffling of neighbourhood structure at the output layer 
that is not specific to the local-global transition itself. We also observe the same pattern in the other models (Appendix~\ref{sec:appendix-crossarch}).

\FloatBarrier

\section{Empirical Verification}
\label{sec:verification}

The theoretical derivation in Section~\ref{sec:transition_point} assumes that contextualised representations of the same word type form local clusters in the embedding space. We test this assumption by measuring the \textbf{same-word fraction}: for each token, the proportion of its $k$ nearest neighbours that share the same word type, evaluated at $k=m-1$ for each lexical diversity condition, and averaged across all datapoints.

\begin{figure}[h!]
    \includegraphics[width=\columnwidth]{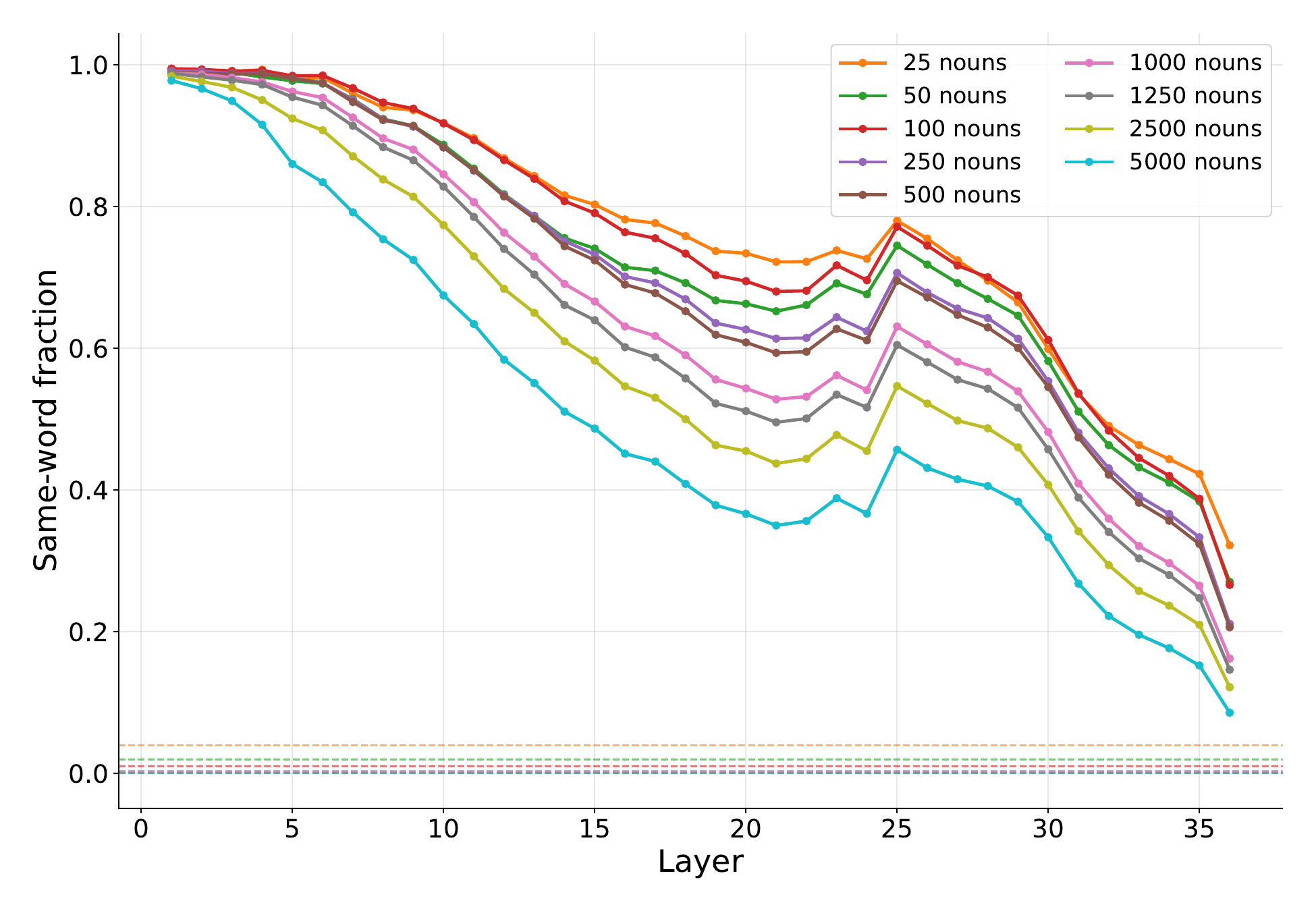}
\caption{Same-word fraction across layers for each lexical diversity condition.}    \label{fig:sameword}
    \end{figure}

Figure~\ref{fig:sameword} shows this fraction across layers for each condition 
(excluding $n=1$, where the fraction is 1, and $n=10{,}000$, where $m-1=0$ makes the diagnostic undefined). In early and middle layers, the same-word 
fraction remains close to 1 for all conditions, confirming that same-word tokens 
almost entirely fill the neighbourhood, exactly as assumed. In later layers, this 
fraction declines, most noticeably for high-diversity conditions, falling to 
around 0.2--0.4 by the final layers. This behaviour is also reflected in the thick transition ID curves in Figure~\ref{fig:gride}, where indeed the last layers display a pronounced peak that is not present otherwise.

The shape of these curves and their final decline show that the local-cloud assumption is not strictly layer-independent: same-word tokens do not remain perfectly segregated from other word types throughout the network. However, even at its lowest point in later layers, the same-word fraction remains far above the no-clustering baseline shown by the dashed lines in Figure~\ref{fig:sameword}.

If representations were not clustered by word type, a token's neighbours would be drawn uniformly from the $n$ word types, giving an expected same-word fraction of approximately $1/n$—three orders of magnitude below the fractions observed even in late layers for the most diverse conditions.

\FloatBarrier

\section{Geometric Properties of the Two Regimes}
Having established that the transition is real and geometrically driven, we 
characterise how ID behaves within each regime. Three consistent observations emerge 
across all scales.

\paragraph{Curve ordering reverses at the transition.}
In the local regime, conditions with fewer unique nouns produce higher ID. With 
fewer unique nouns, more samples per noun are available. We conjecture that, in this setup, the estimator is ``exploring'' to a greater extent the word's contextual variation across all geometric 
directions. The larger the cluster of same-token points around each target word (fewer unique types with more tokens each), the richer this local geometry will be. Hence, we observe ID \textit{decreasing} with higher lexical diversity. In the global regime, on the other hand, the ordering inverts: conditions with more unique nouns produce higher ID. This reversal is visible across all scales in Figure~\ref{fig:gride}. We hypothesise that, as neighbours are now instances of different nouns that will occupy more independent geometric directions in representational space, the larger the set of noun types in the sample, the more accurately we can estimate the overall size of the global noun manifold by looking at a wider set of independent dimensions.

\paragraph{ID saturates in the global regime.}
Within the global regime, ID increases with lexical diversity, but with strongly 
diminishing returns. As $n$ grows large, the global ratio $\rho_{\min} \to 1$ asymptotically: the neighbourhood is already almost entirely filled with different-word tokens, so adding more unique nouns changes its composition only marginally. At scale 512, for instance, the difference in global ratio between 2,500 nouns (99.4\%) and 5,000 nouns (99.8\%) is only 0.4\%, and their ID curves are nearly indistinguishable across all layers, suggesting that, at these sizes, we are reaching a stable estimate of the ID of noun representations.

\paragraph{Do scale effects affect ID profiles?} Arguments about deep net geometry are often cast in terms of the \textit{profile} of ID increase or decrease across network layers. For example, both \citet{valeriani2023geometry} and \citet{cheng2025emergence} reported the presence of an ID ``peak'' in the central layers of LLMs and other models. By looking at the profiles in Figure~\ref{fig:gride}, we see that the central peak pattern is relatively stable across scales and datasets, suggesting that it is a genuine property of how the geometric organisation of representations evolves across network layers. From this perspective, it is indeed interesting that the peak emerges in both regimes, suggesting that it is a property that characterises how representations are organised by the network both within tight, clustered neighbourhoods (local regime), and when considering a wide sample of unrelated points spanning a large space (in our case, the nominal domain, as explored in the global regime). However, note that at the transition point (thicker lines in the same plot), we also observe a second, sharper peak. It would be tempting to interpret this second peak as another substantive property of how models distribute representations in their space, but actually our analysis suggests that it is simply an artefact of the transition from the local to the global regime in the later layers.

\FloatBarrier

\section{Cross-Validation}
\label{sec:crossval}

To test whether the transition reflects a general geometric property of lexical 
diversity, rather than an artefact of our specific estimator, model, or word 
class, we replicate our main analysis along three independent axes: estimator, 
model architecture, and part-of-speech composition. Full results for each are 
given in Appendix~\ref{sec:appendix-crossval}.

\subsection{MLE Replication}
\label{sec:mle-replication}
We repeat the analysis using the Maximum Likelihood Estimator (MLE) of 
\citet{levina2004maximum}, as implemented in the \texttt{skdim} library
\citep{bac2021scikit}, in place of GRIDE. The same transition pattern emerges at 
every scale $k$: the condition at $n = N/k$ nouns separates from the cluster, 
matching the GRIDE results exactly (Appendix~\ref{sec:appendix-crossval}, 
Figure~\ref{fig:mle-appendix}). This indicates the transition is a property of 
the underlying representations rather than a characteristic of GRIDE-based ID estimation specifically.

\subsection{Cross-Architecture Validation}
\label{sec:lama-replication}

We repeat the analysis on Meta-Llama-3-8B and on six further models: Mistral-7B, DeepSeek-LLM-7B, and Pythia at four sizes (410M, 1.4B, 6.9B and 12B). In all models, the transition occurs at the predicted point (Appendix~\ref{sec:appendix-crossarch}), indicating that the transition point is governed by dataset construction rather than by the architecture or size of any single model.

\subsection{POS-diversified Dataset}
Finally, we test whether the transition depends on nouns specifically, or 
reflects lexical diversity more generally. We construct a POS-diversified dataset, sampling words across nouns, verbs,
adjectives, adverbs, and proper nouns, with adverbs excluded at the highest
diversity levels due to data availability
(Appendix~\ref{sec:appendix-allpos}). The transition 
pattern is preserved across all six scales (Appendix~\ref{sec:appendix-crossval}, 
Figure~\ref{fig:balanced-appendix}), indicating that the effect is driven by 
lexical diversity itself rather than any property specific to nouns.

\FloatBarrier
\section{Discussion and Conclusion}
The central finding of this work is that neighbourhood-based ID estimators applied to LLM contextualised word representations operate in two qualitatively distinct modes depending on the relationship 
between lexical diversity, sample size, and scale. In the local regime, they 
measure the contextual variation of individual lexical items. In the global 
regime, they measure how the model distributes different word types relative to 
one another. These are meaningful geometric quantities, but they are not the 
same, and comparing ID values across the two regimes is not 
straightforward.

This point is illustrated directly by our own results. The noun-versus-preposition comparison in Figure~\ref{fig:noun_adp} involved two 
word classes that differ substantially in lexical diversity, and, once they were matched with respect to this property, the result of the comparison reversed. This suggests that the original 
difference was driven by how the underlying dataset was 
constructed, rather than by any intrinsic property of the syntactic categories themselves.

The relationship between lexical diversity and ID likewise differs between the two regimes.
Below the transition, lexical diversity and ID are negatively related: more unique nouns mean fewer samples per noun, sparser cloud sampling, and lower ID. 
Above the transition, they are positively related: more unique nouns mean more 
word clouds, a richer inter-word geometry, and higher ID. At the transition point, spurious ID effects might appear, such as a second ``peak'' in the ID profile that is due to the transition and a more pronounced neighbourhood reorganisation.

We also observed, within the 
global regime, a ceiling beyond which increasing lexical diversity 
yields no additional geometric information. This ceiling is determined by the 
global ratio approaching 1 and can be predicted in advance from 
Equation~\ref{eq:globalratio}. On a positive note, the fact that we observe this ceiling suggests that the sample sizes we are considering are sufficient to come up with a stable estimate of global-regime ID.

More broadly, our findings suggest that ID differences attributed to a 
linguistic or task-relevant property may in some cases reflect uncontrolled 
differences in dataset construction between the compared categories, rather than 
the property itself. Any study using ID to compare word representations across 
conditions with different lexical diversity must verify which regime each 
condition falls into. The formula $n_{\mathrm{transition}} = N/k$ and the 
same-word fraction diagnostic introduced in Section~\ref{sec:verification} 
provide the tools to do this. Conditions in the same regime can be 
compared directly; conditions in different regimes measure different 
aspects of the geometry and require careful interpretation.

Note that we have focused on lexical diversity because it is an easy-to-control property with a predictable effect on neighbourhood structure, but other properties, whose effect might be more difficult to measure, might also affect how data are clustered and, consequently, the ID regime we are in. For example, when comparing a generic text to a highly specialised one (e.g., in a technical domain), it is possible that samples from the latter will form a tighter cluster than samples from the former. Future work should assess the impact of such fuzzier factors on ID.

It is interesting and relevant to underline that ID can be used as a probe of the structure of both local neighbourhoods, where the space surrounding a point is occupied by very similar representations, and global neighbourhoods, such as the large space spanned by a random selection of unrelated nouns. It is remarkable that, while other properties change with scale, certain basic features of ID profiles, such as the presence of a mid-layer peak or bump \citep{valeriani2023geometry,cheng2025emergence}, appear in both regimes, suggesting a ``fractal-like'' structure where the same patterns emerge at narrower and wider sections of the same manifold. Future work should seek to understand why such patterns emerge.

\section*{Limitations}

Our study has several limitations. First, our dataset is derived entirely from 
WikiText-103, an English-language corpus of Wikipedia articles. Whether the 
transition formula $n_{\mathrm{transition}} = N/k$ generalises to other 
languages, genres, or domains with different distributional properties remains 
untested.

Second, while we validate our findings across several architectures and across model sizes from 410M to 12B parameters, we have not tested substantially larger models, and it remains open whether the transition behaves in the same way at larger scales.

Third, our theoretical derivation and the same-word fraction diagnostic in 
Section~\ref{sec:verification} rely on the assumption that contextualised 
representations of the same word type cluster more closely together than 
representations of different words. We find that this assumption holds 
strongly in early and middle layers, but weakens in later layers, where the 
same-word fraction declines substantially for high-diversity conditions. 
Although this decline does not prevent the transition point from matching our 
predictions even in late layers, it indicates that same-word tokens are not 
perfectly segregated from other word types at every layer: same-word 
neighbours dominate far more often than chance would predict, but not with 
absolute certainty, and the reasons for this late-layer weakening are not fully 
understood.
\section*{Acknowledgments}
MB and IM received funding from the European Research Council (ERC) under the European Union's Horizon 2020 research and innovation program (grant agreement No.\ 101019291). We thank the members of the Computational Linguistics and Linguistic Theory (COLT) group at UPF and the ARR reviewers and area chair for their valuable feedback and discussions.

\bibliography{custom}

\appendix

\section{POS-Diversified Dataset}
\label{sec:appendix-allpos}

In addition to the noun-only dataset described in Section~\ref{sec:setup}, we 
construct a POS-diversified variant in which target words are sampled across 
multiple part-of-speech categories. The dataset follows the same construction 
procedure and lexical diversity levels as the main dataset, with $N = 10{,}000$ 
samples per condition.

For $n \leq 1{,}250$ unique words, target words are sampled uniformly across five 
POS categories: \textsc{noun}, \textsc{verb}, \textsc{adj}, \textsc{adv}, and 
\textsc{propn} (proper noun). For $n \geq 2{,}500$, adverbs are excluded and sampling is restricted to four categories: \textsc{noun}, \textsc{verb}, \textsc{adj}, and 
\textsc{propn}. This adjustment is necessary because adverbs occur with insufficient 
frequency in the corpus to satisfy the minimum sample requirements at high lexical diversity levels. The same POS purity and minimum frequency filters applied to the noun-only dataset 
are applied independently to each POS category.

\section{Transition Point by Lexical Diversity}
\label{sec:appendix-transition-by-n}
\begin{figure}[h!]
\centering
\includegraphics[width=0.88\columnwidth]{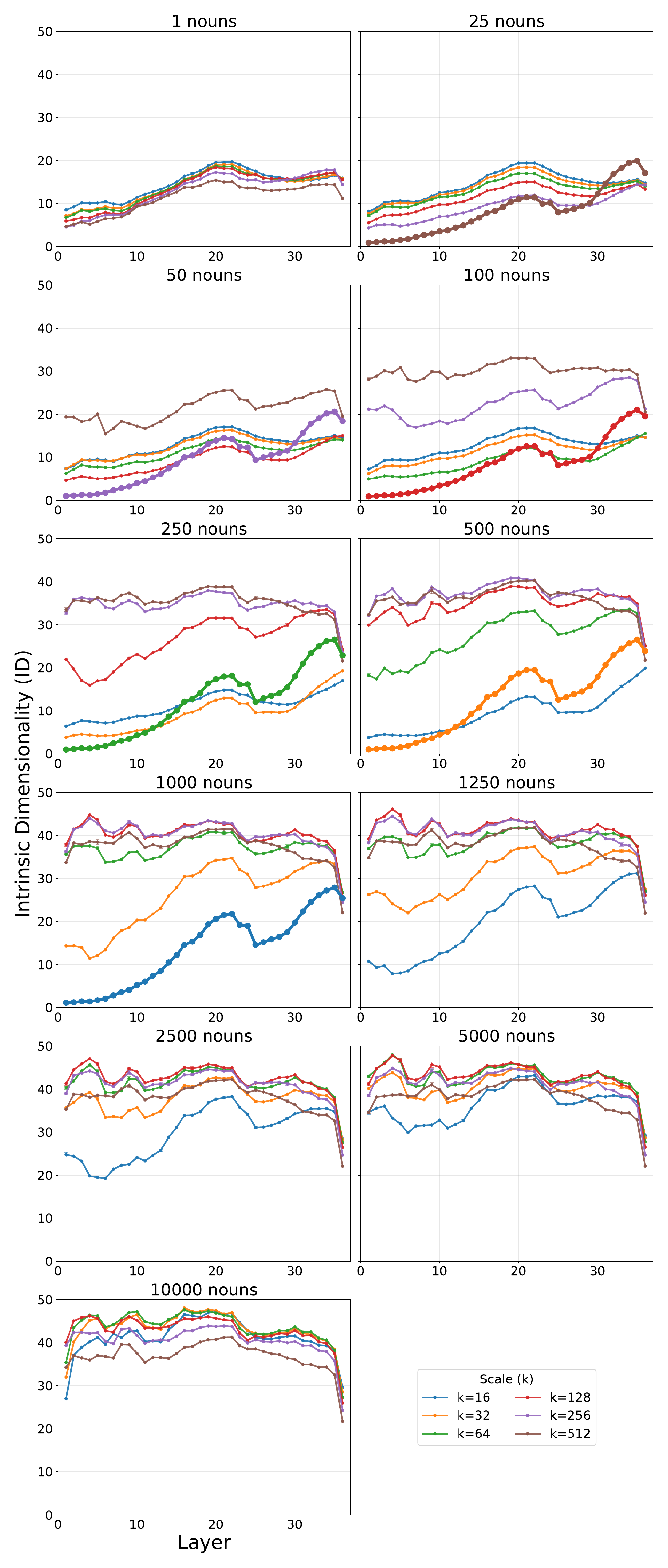}
\caption{ID curves across layers for each lexical diversity level $n$, with one line per scale $k$. The bold line marks the transition scale.}
  \label{fig:transition_by_n}
\end{figure}

Figure~\ref{fig:transition_by_n} presents the same data as 
Figure~\ref{fig:gride}, reorganised to allow a complementary reading: each panel 
now corresponds to a fixed lexical diversity level $n$, with one line per scale 
$k$, rather than a fixed scale with one line per $n$. Within each panel, the 
bolded line marks the scale $k$ at which that lexical diversity level serves as 
the transition point between the local and global regimes. 

\FloatBarrier

\section{Cross-Validation Results}
\label{sec:appendix-crossval}
This appendix presents the full figures supporting the robustness checks 
described in Section~\ref{sec:crossval}: replication with an alternative ID 
estimator (MLE), replication across model architectures, and replication on a 
part-of-speech-diversified dataset.

\subsection{MLE Replication}
\begin{figure}[h!]
\includegraphics[width=\columnwidth]{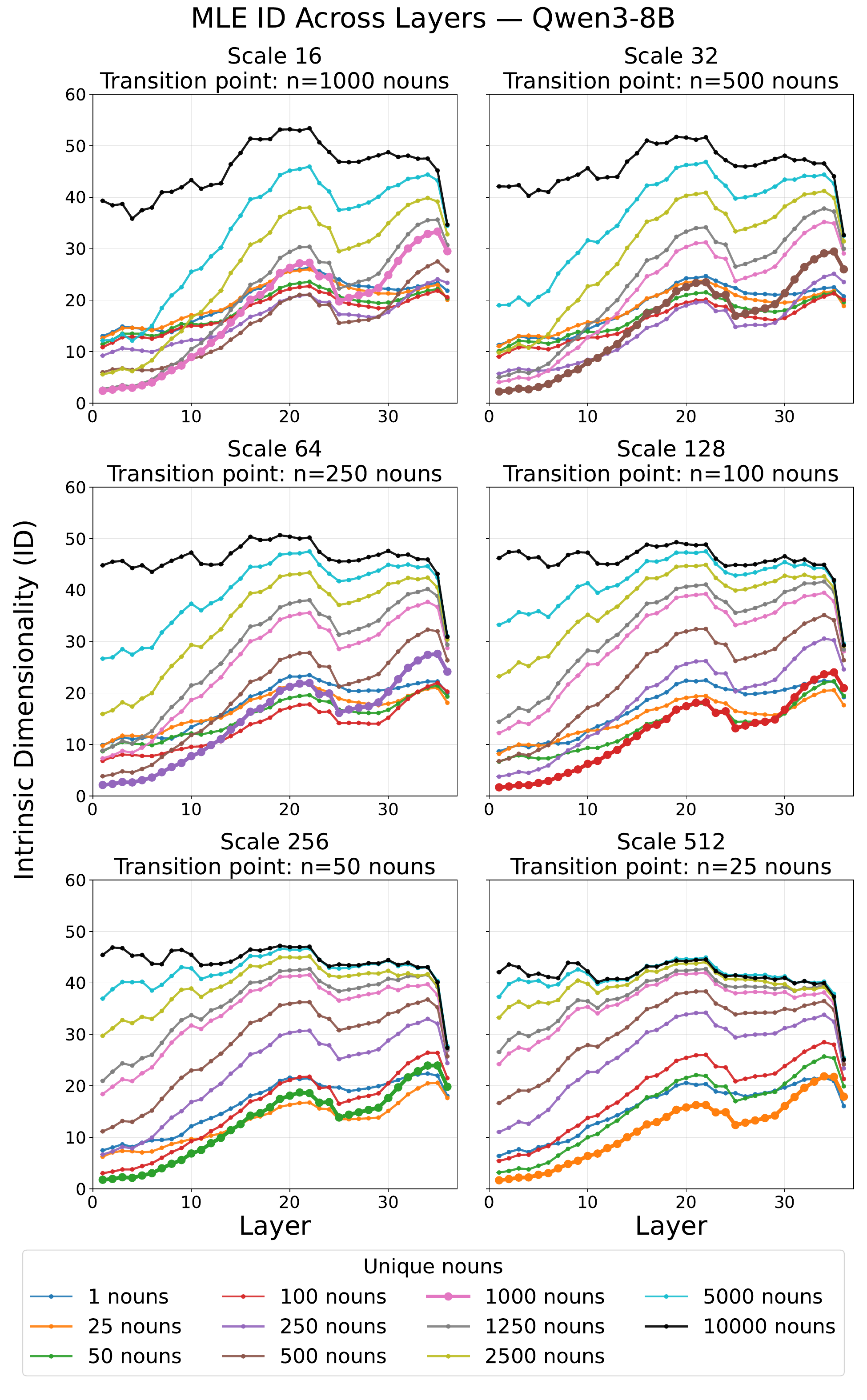}
  \caption{MLE ID curves across layers for all scales on Qwen3-8B. The transition 
  pattern observed with GRIDE (Figure~\ref{fig:gride}) is reproduced at every 
  scale $k$.}
  \label{fig:mle-appendix}
\end{figure}

Figure~\ref{fig:mle-appendix} shows ID curves across all layers of Qwen3-8B at 
all six scales, estimated using the Maximum Likelihood Estimator (MLE) of 
\citet{levina2004maximum} in place of GRIDE. The transition pattern matches the 
GRIDE results in Figure~\ref{fig:gride} exactly, with the condition at $n=N/k$ 
nouns separating from the cluster at every scale.

\subsection{Cross-Architecture Validation}
\label{sec:appendix-crossarch}
\begin{figure}[h!]
\includegraphics[width=\columnwidth]{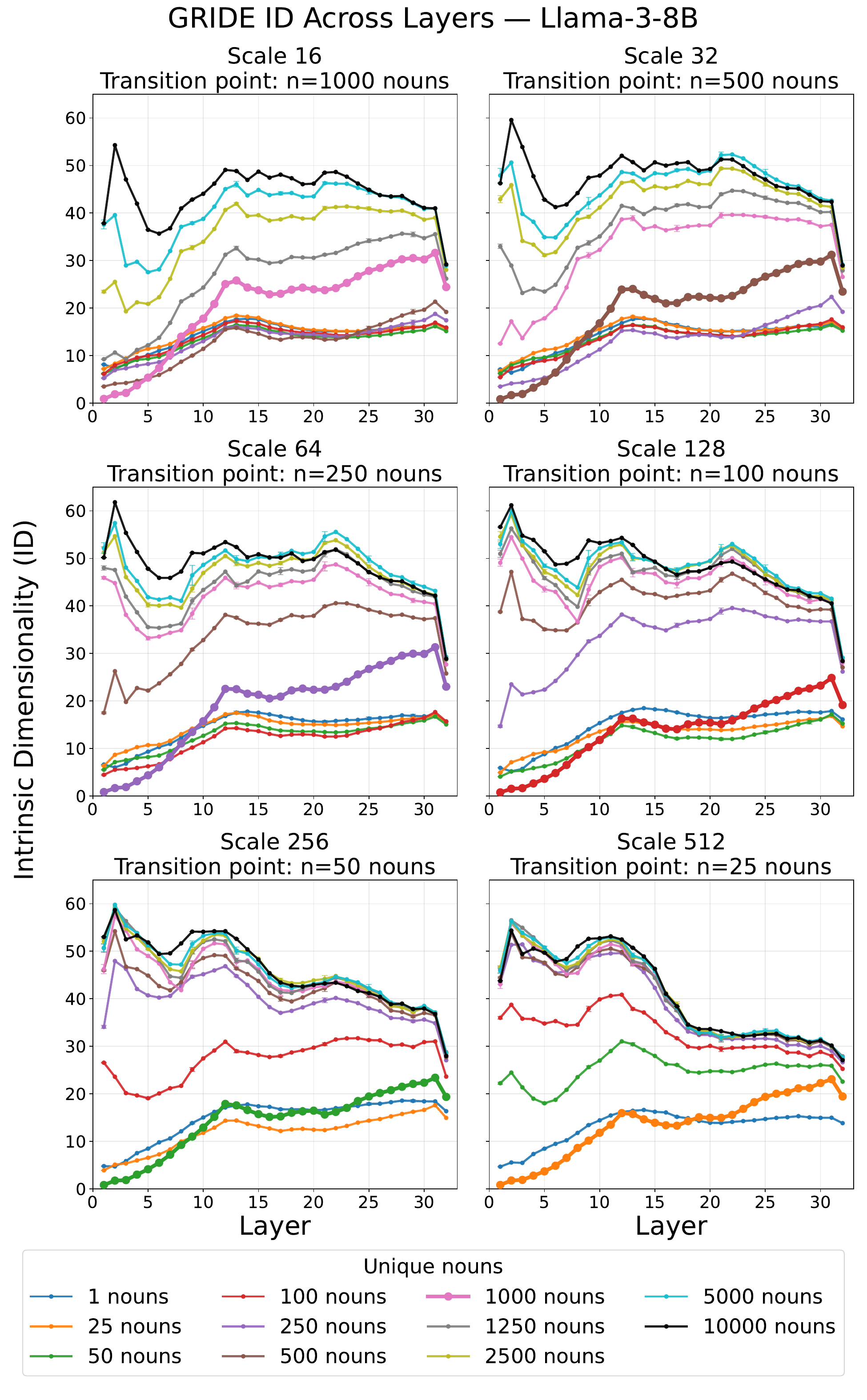}
  \caption{ID curves across all layers for all scales on Meta-Llama-3-8B. The 
  transition occurs at $n = N/k$ nouns at every scale, matching the pattern 
  observed on Qwen3-8B (Figure~\ref{fig:gride}).}
  \label{fig:llama-appendix}
\end{figure}

Figure~\ref{fig:llama-appendix} shows ID curves across all layers and all six 
scales for Meta-Llama-3-8B. The transition occurs at the predicted level 
$n_{\mathrm{transition}} = N/k$ at every scale, matching the pattern observed on 
Qwen3-8B. 

\begin{figure*}[p]
\centering
\includegraphics[width=0.48\textwidth]{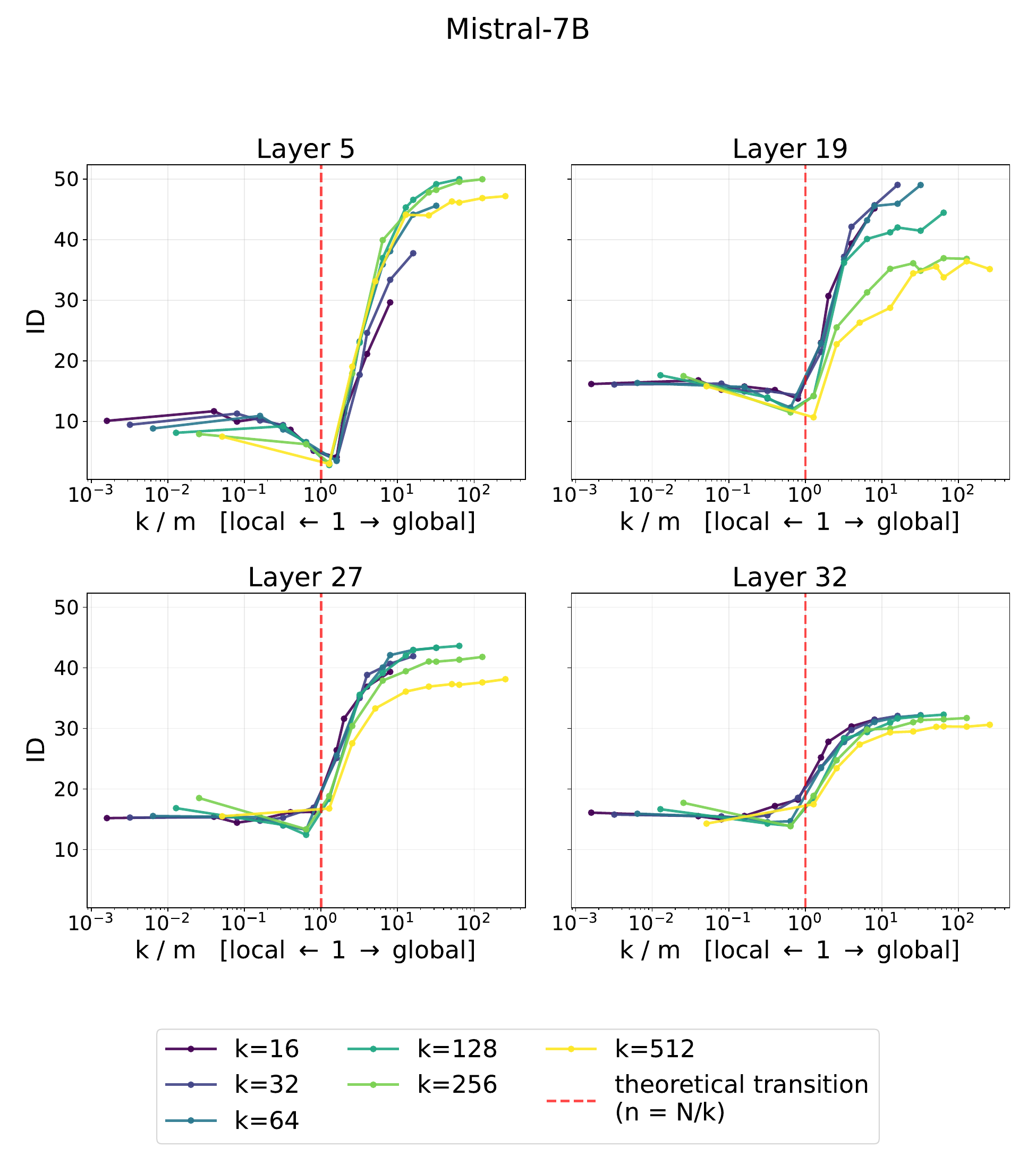}\hfill
\includegraphics[width=0.48\textwidth]{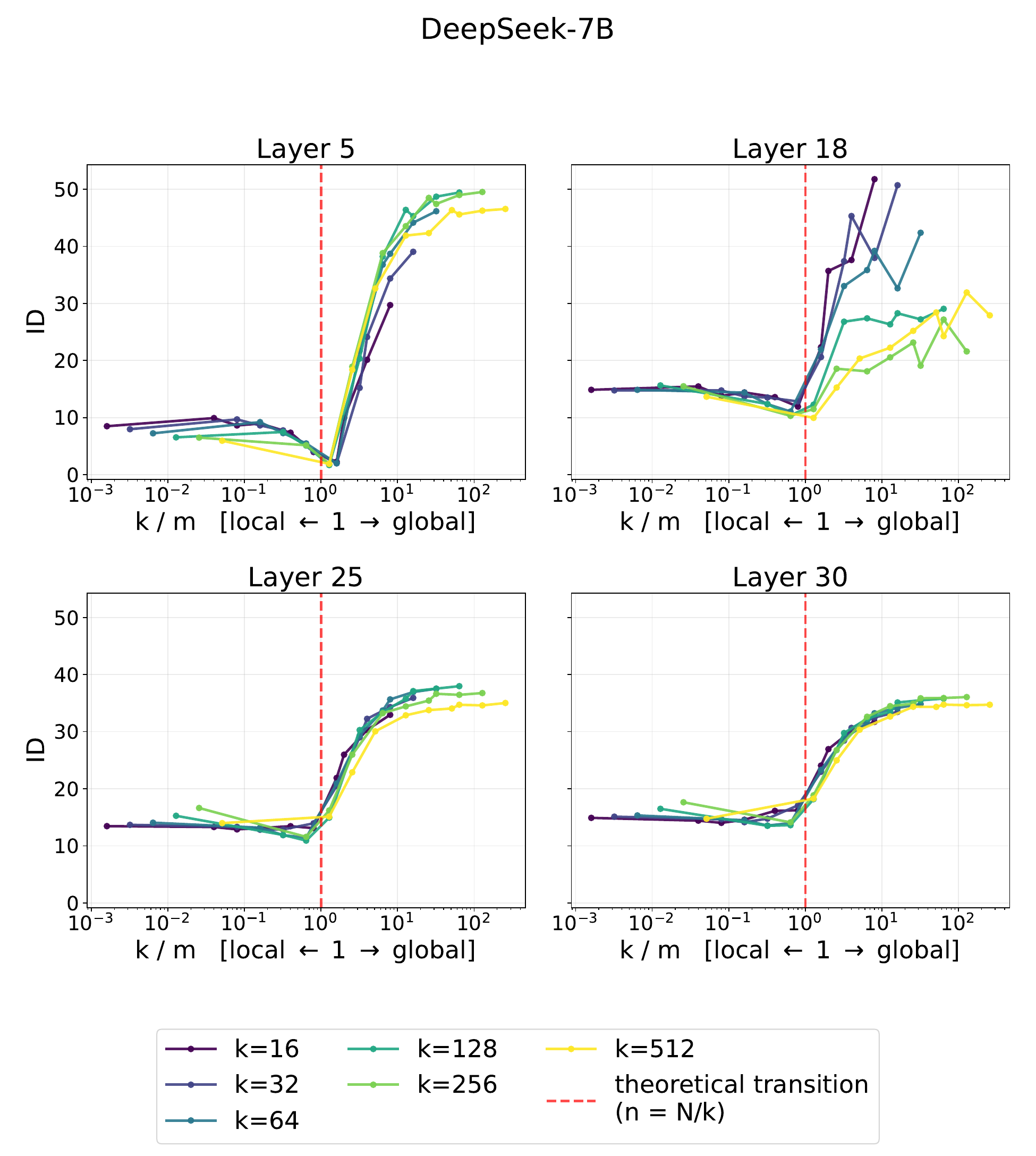}
\caption{ID as a function of $k/m$ for Mistral-7B (left) and DeepSeek-LLM-7B (right), at four layers each.}
\label{fig:more-models}
\end{figure*}

\begin{figure*}[p]
\centering
\includegraphics[width=0.48\textwidth]{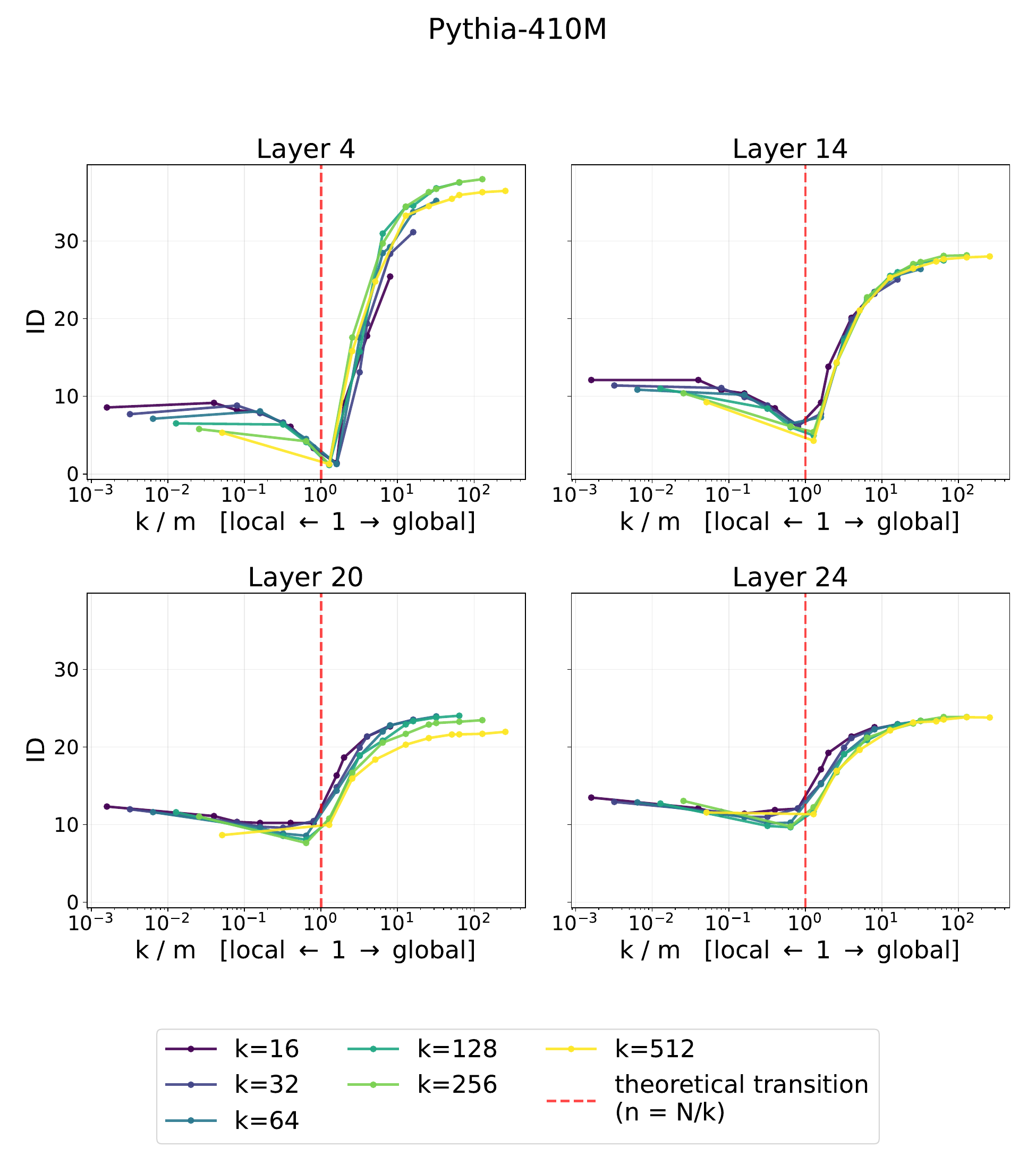}\hfill
\includegraphics[width=0.48\textwidth]{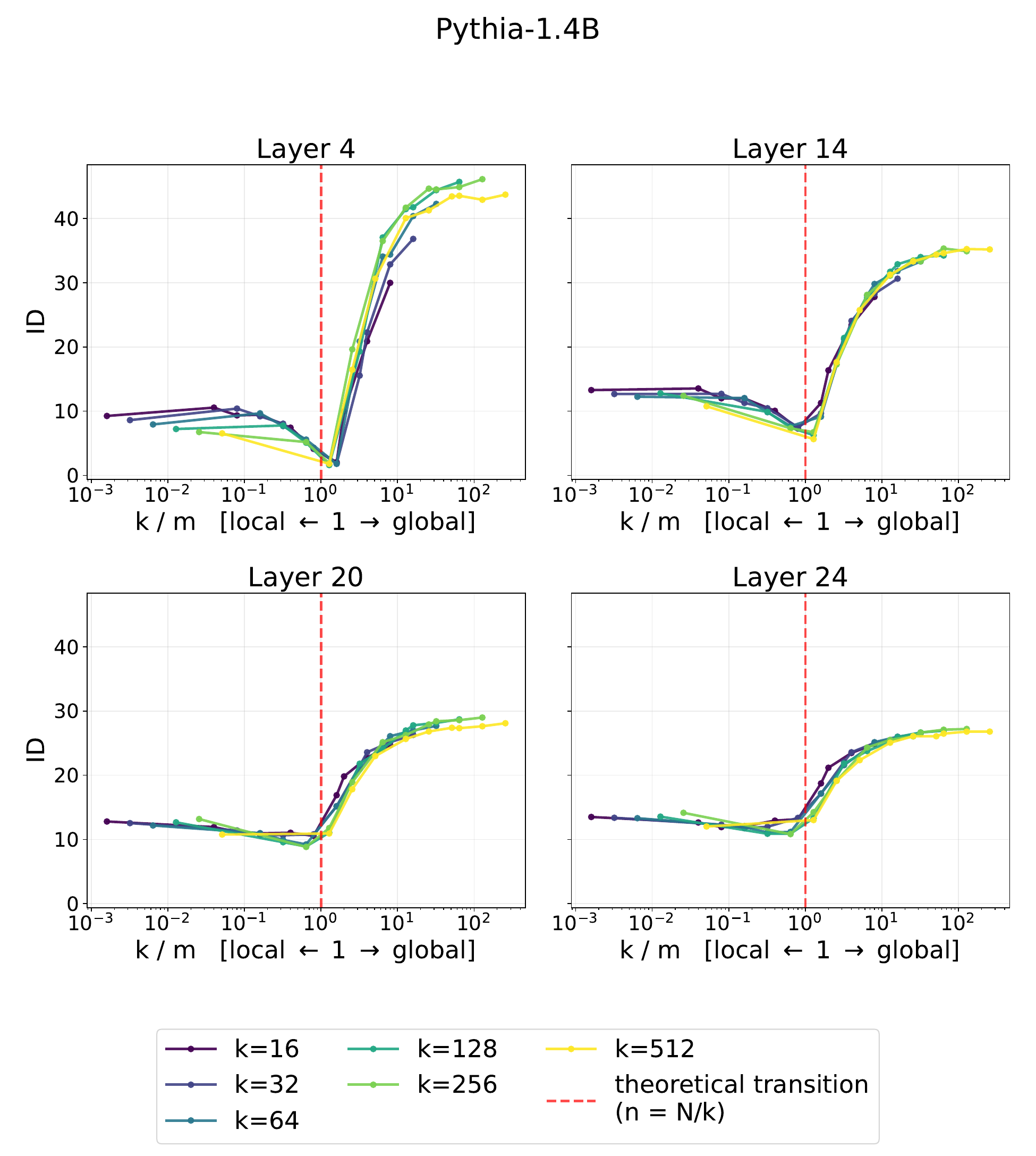}\\[1em]
\includegraphics[width=0.48\textwidth]{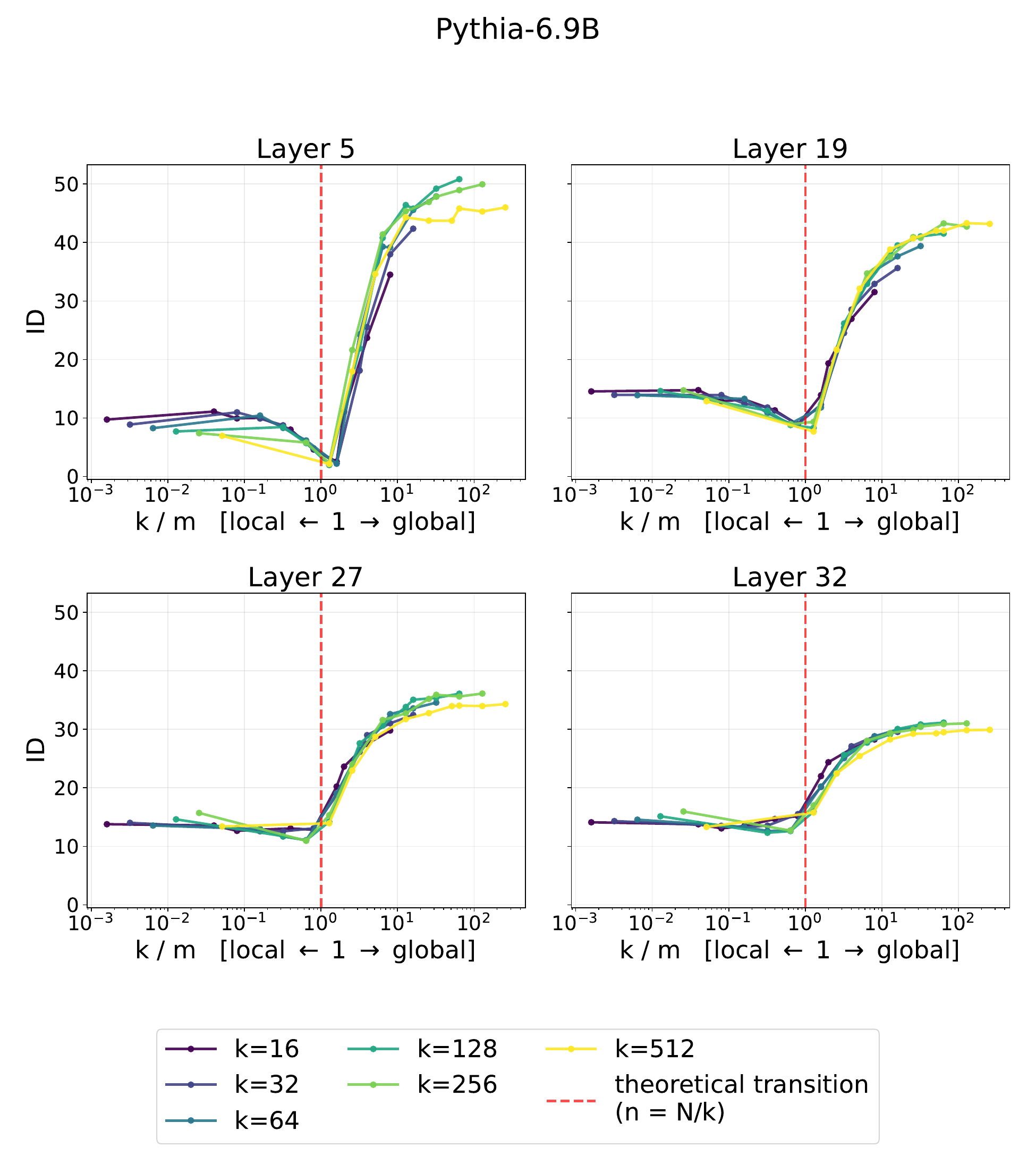}\hfill
\includegraphics[width=0.48\textwidth]{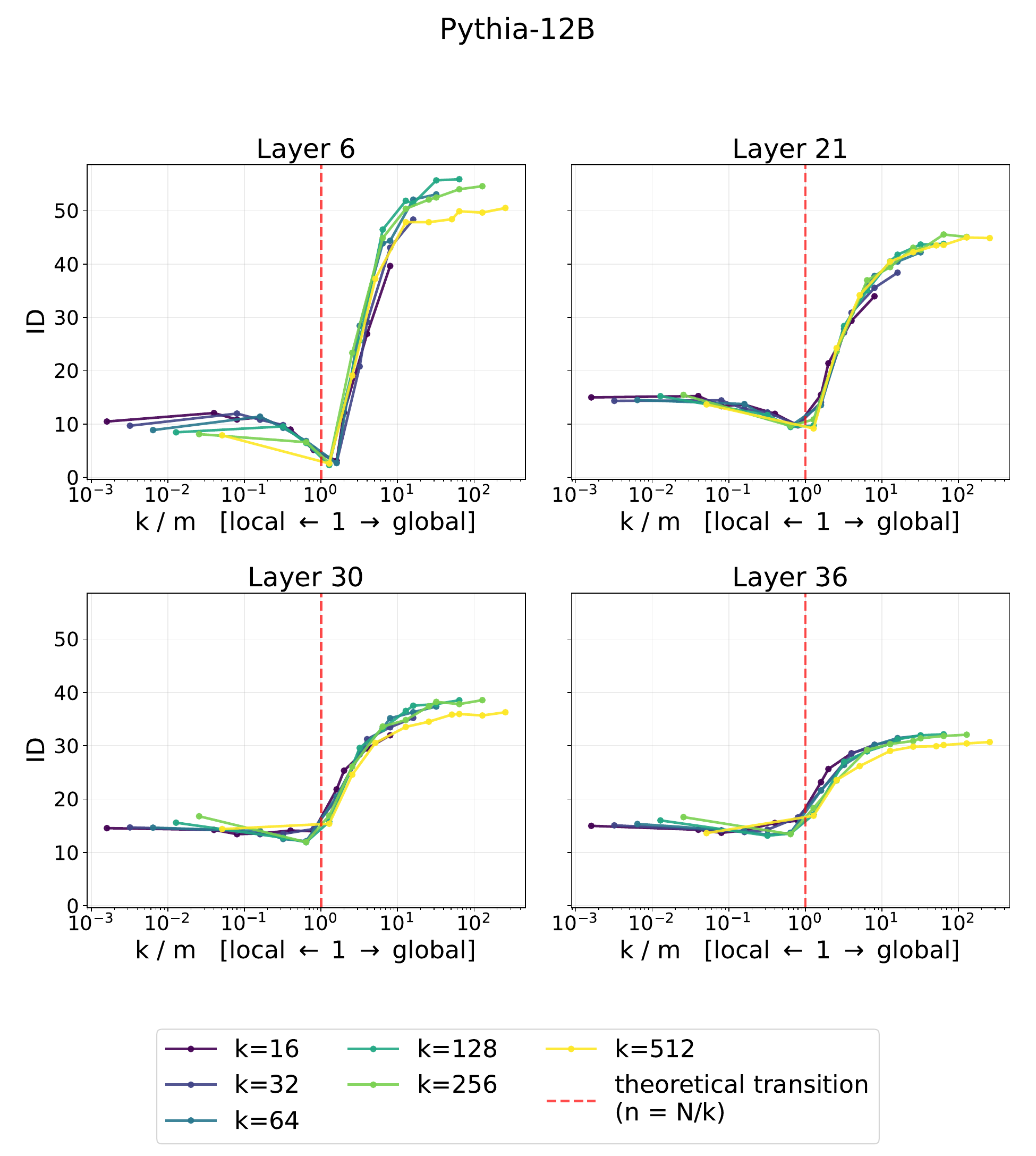}
\caption{ID as a function of $k/m$ for four Pythia sizes: 410M (top left), 1.4B (top right), 6.9B (bottom left) and 12B (bottom right), at four layers each.}
\label{fig:pythia-sizes}
\end{figure*}

Figures~\ref{fig:more-models} and~\ref{fig:pythia-sizes} show the same analysis for six additional models, plotting ID against $k/m$ as in Figure~\ref{fig:collapse}. In every model, the curves switch regime at $k/m = 1$ (dashed red line), as predicted.


\FloatBarrier

\subsection{POS-diversified Dataset}
\begin{figure}[h!]
\includegraphics[width=\columnwidth]{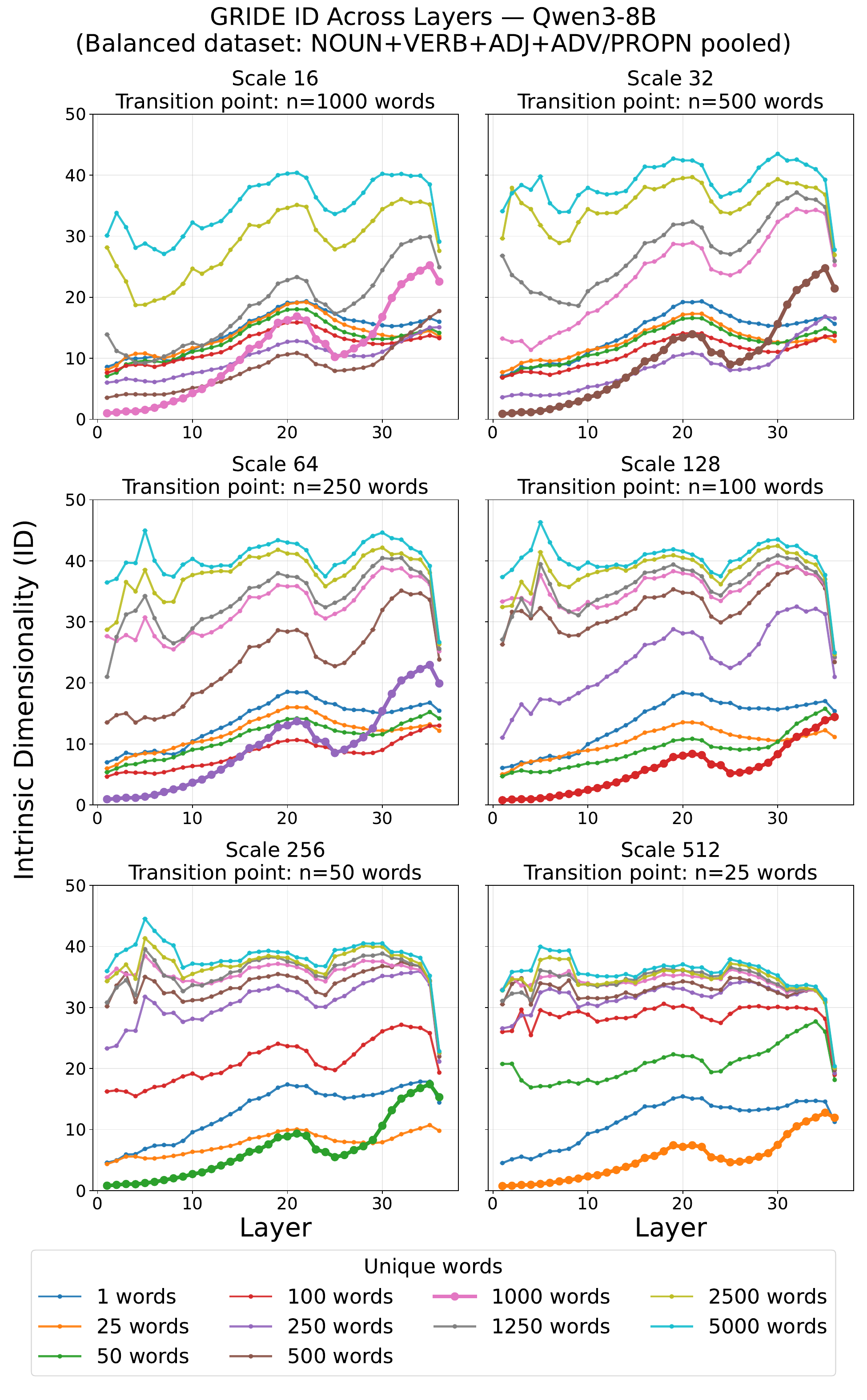}
  \caption{ID curves across layers for all scales on Qwen3-8B, computed on the 
  POS-diversified dataset (Appendix~\ref{sec:appendix-allpos}). The transition 
  pattern is preserved at each scale $k$, matching the noun-only results in 
  Figure~\ref{fig:gride}.}
  \label{fig:balanced-appendix}
\end{figure}

Figure~\ref{fig:balanced-appendix} shows ID curves across layers and all six 
scales for Qwen3-8B, computed on the POS-diversified dataset described in 
Appendix~\ref{sec:appendix-allpos}. The transition pattern is preserved at every 
scale, indicating that the effect is driven by lexical diversity itself rather 
than by any property specific to nouns.

\FloatBarrier
\section{Licenses}
\label{sec:appendix-licenses}
URLs and licenses of the used assets are provided in the following list:

\vspace{0.5em}
\noindent\textbf{WikiText-103} \url{https://huggingface.co/datasets/Salesforce/wikitext}; 
license: CC BY-SA 3.0

\vspace{0.5em}
\noindent\textbf{spaCy} \url{https://spacy.io}; 
license: MIT

\vspace{0.5em}
\noindent\textbf{Qwen3-8B} \url{https://huggingface.co/Qwen/Qwen3-8B}; 
license: apache-2.0

\vspace{0.5em}
\noindent\textbf{Llama} \url{https://huggingface.co/meta-llama/Meta-Llama-3-8B}; 
license: llama3

\vspace{0.5em}
\noindent\textbf{DADApy} \url{https://github.com/sissa-data-science/DADApy}; 
license: apache-2.0

\vspace{0.5em}
\noindent\textbf{skdim} \url{https://github.com/scikit-learn-contrib/scikit-dimension}; 
license: BSD-3-Clause

\vspace{0.5em}
\noindent\textbf{Mistral-7B} \url{https://huggingface.co/mistralai/Mistral-7B-v0.3}; 
license: apache-2.0

\vspace{0.5em}
\noindent\textbf{DeepSeek-LLM-7B} \url{https://huggingface.co/deepseek-ai/deepseek-llm-7b-base}; 
license: DeepSeek License

\vspace{0.5em}
\noindent\textbf{Pythia} \url{https://huggingface.co/EleutherAI}; 
license: apache-2.0

\end{document}